\documentclass{article}
\usepackage{spconf,amsmath,graphicx}
\usepackage{amsmath, amssymb}
\usepackage{pifont}
\usepackage{booktabs}
\usepackage{multirow}
\usepackage{hyperref}
\usepackage[table,xcdraw]{xcolor}
\usepackage{enumitem}
\setlist{nosep, leftmargin=14pt}
\usepackage{mwe} 
\usepackage{float}

\title{MedReason-R1: Learning to Reason for CT Diagnosis with Reinforcement Learning and Local Zoom}
\name{Yifan Li$^{1,2,3}$ \qquad Fenghe Tang$^{1,2,3}$ \qquad Yingtai Li$^{1,2,3}$ \qquad Shaohua Kevin Zhou$^{1,2,3,4 \star}$}
\address{$^1$ School of Biomedical Engineering, Division of Life Sciences and Medicine, \\ University of Science and Technology of China, Hefei, Anhui, 230026, P.R. China\\
$^{2}$ Center for Medical Imaging, Robotics, and Analytic Computing \& LEarning (MIRACLE), \\
Suzhou Institute for Advanced Research, USTC, Suzhou 215123, P.R. China \\
$^{3}$ Jiangsu Provincial Key Laboratory of Multimodal Digital Twin Technology, Suzhou Jiangsu, 215123 \\
$^{4}$ State Key Laboratory of Precision and Intelligent Chemistry, USTC, Hefei Anhui 230026, P.R. China
}

\begin{document}
%
\maketitle

\begin{figure*}[htbp]
  \centering
  \includegraphics[width=1\textwidth]{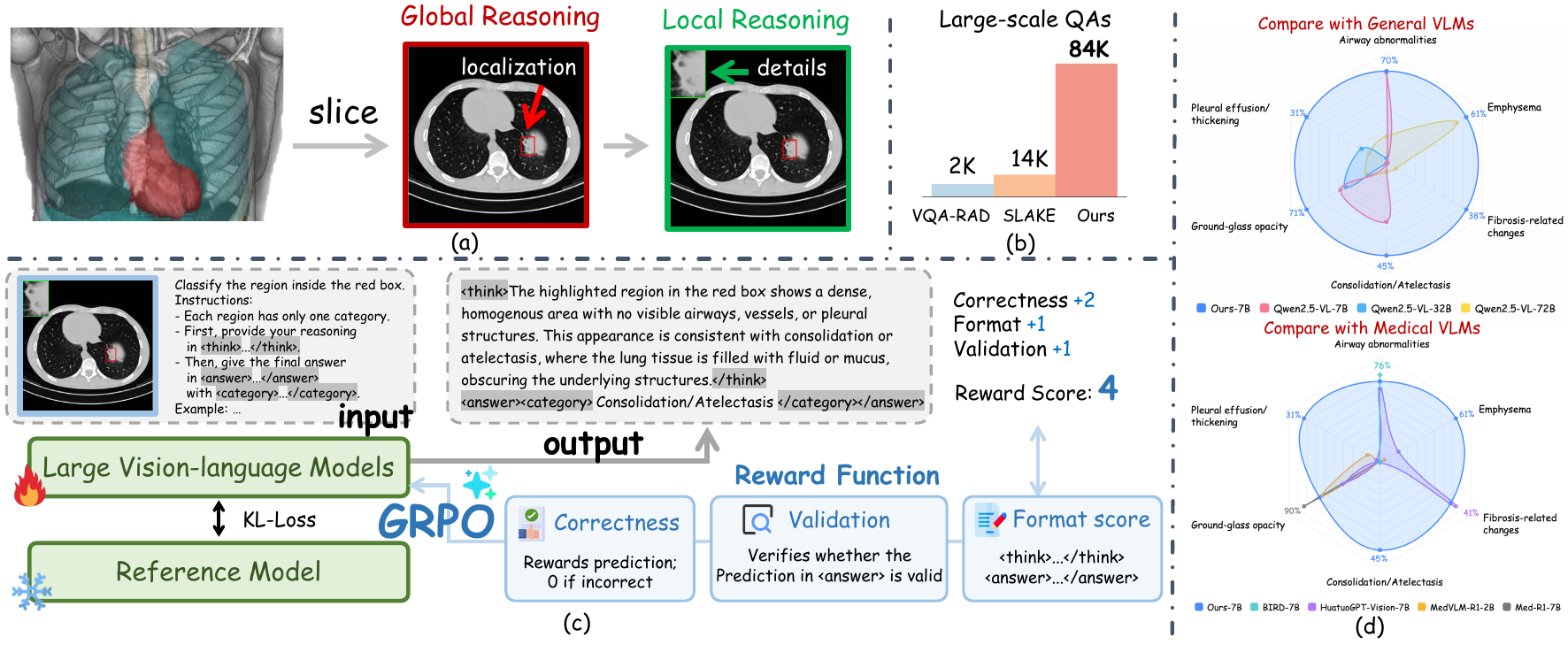}
  \vspace{-8mm}
  \caption{Framework and results of the proposed method. 
  (a) Global-to-local reasoning enables lesion localization and fine-grained feature analysis.\label{fig:grpo_a}
  (b) Our dataset comprises 84K QA pairs, substantially larger than existing VQA-RAD and SLAKE.\label{fig:grpo_b}
  (c) GRPO pipeline integrates correctness, format, and validation rewards to optimize model outputs.\label{fig:grpo_c}
  (d) Comparative evaluation shows our model achieves superior or competitive performance against both general-purpose and medical vision–language models on chest CT abnormalities.\label{fig:grpo_d}}
  \vspace{-3mm}
  \label{fig:grpo_architecture}
\end{figure*}

\begin{abstract}
General-purpose large Vision-Language Models (VLMs) demonstrate strong capabilities in generating detailed descriptions for natural images. However, their performance in the medical domain remains suboptimal, even for relatively straightforward tasks, primarily due to the lack of large-scale, high-quality, specialized medical imaging datasets and the neglect of the diagnostic process that progresses from coarse to fine-grained. To address the first issue, we construct the CT-RATE-VQA dataset, which has 84K QA pairs. 
For the second issue, we propose MedReason-R1, a medical VLM with explicit reasoning process for disease diagnosis. MedReason-R1 incorporates a novel 
strategy that embeds zoom-in disease region-of-interest areas into the image, highlighting the crucial role of both global localization and disease-specific details in enhancing the model's diagnostic performance. Furthermore, we introduce the GRPO reinforcement learning framework to MedReason-R1, which enables effective reasoning without relying on costly manual annotations. 
Compared to recent general-purpose and medical VLMs, MedReason-R1 achieves state-of-the-art performance in CT disease diagnosis while retaining generalization. The code, checkpoints, and  dataset are available at  \href{https://github.com/Leevan001/MedReason-R1}{\textcolor{magenta}{\textit{this URL}}}.
\end{abstract}
\begin{keywords}
Med-VQA, Multimodal Large Language Models (MLLMs), Reinforcement Learning
\end{keywords}
\section{Introduction}
\label{sec:intro}
Medical Visual Question Answering (Med-VQA) ~\cite{Llava-med,bird,chen2022align,gong2021cross} is to generate natural language answers from medical images to assist clinicians in diagnosis and decision-making, and to enhance model interpretability for trustworthy clinical deployment. Efficient Med-VQA systems can help physicians rapidly interpret imaging features, improve diagnostic efficiency, and provide patients with personalized medical consultation.
With the development of Multimodal Large Language Models (MLLMs)~\cite{qwen,yao2024minicpm,zhu2025internvl3}, general-purpose models have demonstrated strong capabilities in natural image understanding and text generation. However, even for relatively simple CT disease classification tasks, existing models often underperform~\cite{liu2024survey}. This limitation primarily stems from three factors: \textbf{(1) Limited Med-VQA Datasets.} The lack of large-scale, high-quality, specialized medical imaging datasets restricts the training and performance of models in the medical domain. \textbf{(2) Difficulty in Capturing Disease Features.} The blurred disease boundaries and scale variability pose a challenge for VLMs to capture both global and local lesion features. \textbf{(3) Absence of Accurate Reasoning Steps}, resulting in outputs with limited interpretability.

To address the aforementioned challenges, this study introduces improvements from three perspectives: data construction, feature modeling, and reasoning capability.
First, to mitigate the limitation of existing Med-VQA datasets, we construct a large CT-based visual question answering dataset (in Fig.~\ref{fig:grpo_architecture}(b)), CT-RATE-VQA, which contains 84k rich lesion-level training samples for model learning.
Second, to address the difficulty of visual-language models in both localization and capturing fine-grained lesion details, we propose a local lesion augmentation strategy that instructs VLMs to simulate the physician's diagnostic process: first, accurately locate the disease, then focus on the disease's detailed characteristics (in Fig.~\ref{fig:grpo_architecture}(a), the red box for localization and the green box for lesion details).
Finally, to overcome the lack of explicit reasoning processes in existing models, we incorporate the Group Relative Policy Optimization (GRPO) framework(in Fig.~\ref{fig:grpo_architecture}(c)), which employs structured reward functions to guide the model in autonomously learning effective visual reasoning strategies without relying on manually annotated chains of thought, thereby enhancing both reasoning capability and interpretability~\cite{guo2025deepseek,shao2024deepseekmath,segzero,medr1,medvlm}.

In this work, we propose a comprehensive Med-VQA framework for CT images (Fig.~\ref{fig:grpo_architecture}), which integrates three key components: 
(1) construction of the CT-RATE-VQA dataset with over 84,000 QA pairs; 
(2) design of a local zoom-in patch augmentation method to highlight lesion features (Fig.~\ref{fig:grpo_architecture}(a)); and 
(3) incorporation of task-specific reward functions under the GRPO reinforcement learning framework to enhance the model’s reasoning ability (Fig.~\ref{fig:grpo_architecture}(c)). 

These designs enable the model to more effectively integrate local and global information during reasoning, achieving both stronger diagnostic accuracy (Fig.~\ref{fig:grpo_d}(d)) and better interpretability. Compared with the baseline model, our approach substantially enhances diagnostic performance while preserving generalization.

\section{METHOD}
\label{sec:format}
\subsection{Dataset Construction}
\label{sec:dataset_construction}
We construct a large-scale CT-VQA dataset based on the original ReXGroundingCT data~\cite{rexct} to support model training and evaluation. For each case, CT volumes are processed along with their corresponding multi-class segmentation masks, where each mask channel corresponds to a specific lesion type. At the slice level, slices containing target lesions are selected, and lesion bounding boxes are extracted using connected component analysis. To ensure data quality and balance, we retain boxes larger than 1300 pixels (approximately 0.5\% of the image) and cap each case at 20 uniformly sampled slices. The resulting dataset (Fig.~\ref{fig:grpo_b}(b)) surpasses prior medical VQA data in scale, supporting effective training and evaluation.

To guide VLM reasoning like radiologists, we propose a data augmentation strategy that embeds a zoom-in patch of the lesion in the upper-left corner of each slice (Fig.~\ref{fig:grpo_a}(a)), inspired by the practice of radiologists magnifying regions of interest (ROIs) to examine details. Lesion regions are cropped, resized to a fixed resolution, highlighted with a colored border, and then pasted back onto the original slice. This approach preserves the global anatomical context while emphasizing local lesion features, aiding the VLM in learning discriminative patterns.

\subsection{Policy Optimization with GRPO}

\noindent\textbf{Preliminaries.}
Group Relative Policy Optimization, abbreviated as GRPO~\cite{shao2024deepseekmath}, is a reinforcement learning method that efficiently optimizes policies and performs well in models like DeepSeek R1-Zero~\cite{guo2025deepseek}. Specifically, for a given query $q$, GRPO does not require a critic model like Proximal Policy Optimization (PPO)~\cite{ppo}. Instead, it samples $N$ responses $\{o^{(1)}, o^{(2)}, \allowbreak \dots, \allowbreak o^{(N)}\}$
 from the old policy $\pi_{\theta_\text{old}}$ and uses carefully designed, verifiable task-specific rewards $\{r^{(1)}, r^{(2)}, \dots, r^{(N)}\}$ to estimate the relative advantage of each response $\hat{A}^{(i)}$, enabling direct policy updates.

\begin{equation}
\resizebox{.85\columnwidth}{!}{%
$\begin{split}
\mathcal{J}(\theta) &= \mathbb{E}_{q \sim Q,\, o^{(i)} \sim \pi_{\theta_\mathrm{old}}} \Big[
\min(\rho^{(i)} \hat{A}^{(i)},\, \mathrm{clip}(\rho^{(i)}, 1-\delta, 1+\delta)\hat{A}^{(i)}) \\
&\quad - \beta\, \mathbb{D}_\mathrm{KL}[\pi_{\theta_\mathrm{new}} || \pi_\mathrm{ref}] \Big]
\end{split}$%
}
\end{equation}

Here, $\rho^{(i)} = \pi_{\theta_\mathrm{new}}(o^{(i)}|q)/\pi_{\theta_\mathrm{old}}(o^{(i)}|q)$ is the probability ratio, $\delta$ controls clipping to prevent overly large updates, and $\beta$ weights the KL-divergence regularization. This formulation allows GRPO to integrate strong reward signals while ensuring stable and consistent policy updates.

\noindent\textbf{Reward Design.}
In medical imaging tasks, a model must not only predict the correct disease category but also satisfy strict format and domain-specific constraints. To address this, we design a composite reward function within the GRPO framework, defined as:
\begin{equation}
R = \alpha \cdot R_{\text{format}} + \beta \cdot R_{\text{validity}} + \gamma \cdot R_{\text{correctness}}
\end{equation}
where:
\begin{itemize}
\item $R_{\text{format}}$ evaluates whether the model output follows the prescribed structure, i.e., containing both \texttt{<think>...\allowbreak</think>} and \verb|<answer>...</answer>| sections. This encourages the model to generate structured reasoning, with the final answer enclosed within the \verb|<answer>| tags.
\item $R_{\text{validity}}$ checks whether the predicted category inside \verb|<answer>| belongs to the set of valid disease classes.
\item $R_{\text{correctness}}$ measures the agreement between the predicted and ground-truth categories, assigning a reward for correct predictions and zero otherwise.
\end{itemize}
The hyperparameters $\alpha, \beta, \gamma$ control the relative importance of each reward component. In this study, we set $\alpha = 1$, $\beta = 1.0$, and $\gamma = 2.0$ to prioritize prediction correctness while still encouraging structured output and category validity.
This reward design allows the model, during GRPO-based optimization, to gradually learn to produce structured, medically valid, and accurate diagnostic outputs, thereby enhancing overall reasoning capability and task performance.The whole architecture is shown in Fig.~\ref{fig:grpo_a}(c).
\section{EXPERIMENTS}
\label{sec:pagestyle}
\subsection{Dataset}
We use the CT-RATE-VQA dataset constructed in Section~\ref{sec:dataset_construction}. The dataset is split into training and testing subsets with an 8:2 ratio, consisting of 68,581 training slices and 15,507 testing slices from 3,042 volumes. It covers seven clinically common disease categories (airway abnormalities, emphysema (centrilobular, paraseptal, bullous), fibrosis-related changes, pulmonary nodules/masses, consolidation/atelectasis, ground-glass opacity, and pleural effusion/thickening) with strict patient-level separation.

Compared with existing datasets, VQA-RAD~\cite{VQA-RAD} and SLAKE~\cite{liu2021slake} have a limited number of samples, making them insufficient for training large models. While M3D-VQA~\cite{xin2025med3dvlm} and 3D-RAD~\cite{gai20253d} provide large-scale CT data, they mainly focus on 3D volume-level tasks and have limited or unsystematic slice-level annotations. Our dataset offers large-scale, high-quality slice-level annotations, with categories aligned with clinically common lesions, and is suitable for both slice-level classification and question-answering tasks.

\begin{figure}[t]
  \centering
  \includegraphics[width=0.9\linewidth]{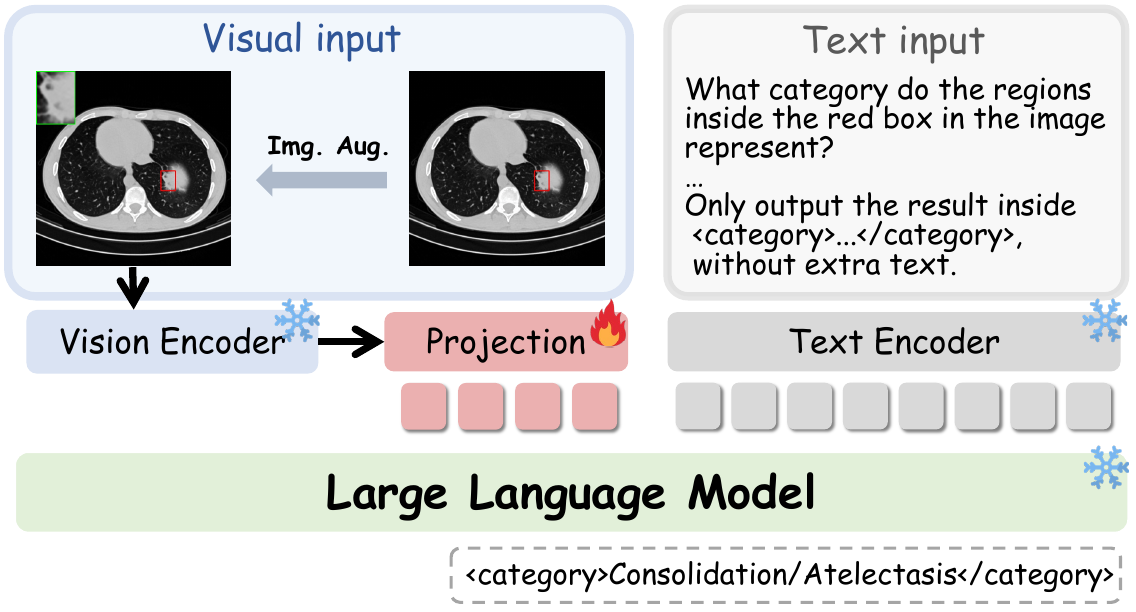}
  \vspace{-3mm}
  \caption{Overview of Stage I: supervised fine-tuning. 
  Only the MLP layers of Qwen2.5-VL are updated to align the model 
  with the disease diagnosis task.}
  \vspace{-3mm}
  \label{fig:sft}
\end{figure}

\subsection{Implementation Details}
\noindent\textbf{Stage I: Supervised Fine-tuning.} As shown in Fig.~\ref{fig:sft}, we initialize our framework by fine-tuning Qwen2.5-VL~\cite{qwen} to medical domains. In this stage, only the MLP layers are updated, following the practice of LLaVA-Med~\cite{Llava-med}. The model is trained for 2 epochs with an initial learning rate of $1\times10^{-5}$, cosine decay, and a warmup ratio of 0.1. All experiments are conducted on 8 H20 GPUs.

\noindent\textbf{Stage II: Reinforcement Learning.} We further optimize the model using the proposed GRPO-based reinforcement learning framework. We set $N=5$ and train for 1,500 steps. This stage enables the model to directly learn from task-specific rewards and refine its reasoning and prediction capabilities.

\subsection{Results Comparison}
We compare our model with state-of-the-art open-source general-purpose and medical-domain MLLMs. To ensure fairness, BiRD~\cite{bird}, initially fine-tuned on Qwen2-VL, was re-trained on Qwen2.5-VL~\cite{qwen} with identical settings. Table~\ref{tab:baseline_summary} summarizes model sizes and average accuracy on CT-RATE-VQA, while Fig.~\ref{fig:grpo_d}(d) shows per-class results. Our model attains balanced performance across all categories, achieving an average accuracy of 0.5218, substantially surpassing comparable 7B models such as Qwen2.5-VL (23.86\%) and HuatuoGPT-Vision (26.65\%). On challenging classes, e.g., Emphysema (60.52\%) and Fibrosis-related changes (38.23\%), it further demonstrates superior generalization. Notably, larger size does not imply stronger performance, as Qwen2.5-VL 32B yields only 13.65\%. BiRD~\cite{bird} exhibits weak instruction-following capability and fails to produce outputs in the required reasoning format, hence results with reasoning are not reported. Overall, while baselines often show biased or limited predictions, our model delivers more balanced and robust recognition of thoracic abnormalities.

\begin{table}[htb]
\centering
\begin{tabular}{lccccc}
\hline
\multirow{2}{*}{\textbf{Model}} & \multirow{2}{*}{\textbf{Size}} & \multicolumn{2}{c}{\textbf{ Accuracy (\%)}} \\
\cline{3-4}
 &  & \textbf{w think} & \textbf{w/o think} \\
\hline
\multicolumn{2}{l}{\textit{\textcolor{gray}{General VLMs}}} & & \\
Qwen2.5-VL\cite{qwen} & 7B  & 22.38 & 23.86 \\
Qwen2.5-VL\cite{qwen} & 32B & 25.43 & 13.65 \\
Qwen2.5-VL\cite{qwen} & 72B & 21.42 & 16.26 \\
\hline
\multicolumn{2}{l}{\textit{\textcolor{gray}{Medical VLMs}}} & & \\
BiRD\cite{bird}       & 7B  &  --   & 7.44 \\
HuatuoGPT-Vision\cite{chen2024huatuogpt} & 7B & 17.57 & 26.65 \\
MedVLM-R1\cite{medvlm}  & 2B  & 20.98 & 20.26 \\
Med-R1\cite{medr1}     & 7B  & 24.31 & 24.64 \\
\rowcolor{gray!15} MedReason-R1 (Ours)       & 7B  & \textbf{52.18} & \textbf{51.68} \\
\hline
\end{tabular}
\vspace{-3mm}
\caption{Comparison of accuracies with and without reasoning on CT-RATE-VQA.}
\label{tab:baseline_summary}
\vspace{-3mm}
\end{table}

\subsection{Impact on General-Domain VQA Performance}
We evaluate our fine-tuned model, which is trained with Supervised Fine-Tuning (SFT) followed by Reinforcement Learning (RL), on ChartVQA~\cite{masry2022chartqa} and TextVQA~\cite{textqa}, where the former assesses chart understanding and the latter scene text recognition. The evaluation metric is accuracy for each dataset, and a weighted average is computed based on the number of samples to provide an overall measure of performance across both tasks. As shown in Table~\ref{tab:vqa_comparison}, the weighted accuracy remains nearly unchanged after SFT and RL (86.98\% vs. 86.87\%), with dataset-specific results of ChartVQA: 78.28\% vs. 77.86\% and TextVQA: 90.35\% vs. 90.32\%. These results indicate that, compared to other models, our approach enhances medical diagnosis capabilities while maintaining stable performance on general VQA tasks.
\begin{table}[h!]
\centering
\resizebox{1\linewidth}{!}
{
\begin{tabular}{lccl}
\toprule
\textbf{Model} & \textbf{ChartVQA} & \textbf{TextVQA} & \textbf{Weighted Avg.} \\
\midrule
\textcolor{gray}{Qwen2.5-VL}\cite{qwen} & \textcolor{gray}{78.3} & \textcolor{gray}{90.3} & \textcolor{gray}{87.0} \\
\rowcolor{gray!15} MedReason-R1 (Ours)  & 77.9 & \textbf{90.3} & \textbf{86.9 $\downarrow$ 0.1} \\
Med-R1\cite{medr1} & 68.6 & 82.2 & 78.4 $\downarrow$ 8.6 \\
BiRD\cite{bird} & \textbf{81.6} & 86.6 & 85.1 $\downarrow$ 1.9 \\
MedVLM-R1\cite{medvlm} & 70.0 & 83.8 & 80.0 $\downarrow$ 7.0 \\
\bottomrule
\end{tabular}
}
\vspace{-3mm}
\caption{Performance comparison on ChartVQA and TextVQA datasets (weighted average reported).}
\label{tab:vqa_comparison}
\vspace{-3mm}
\end{table}


\subsection{Ablation Study}
\noindent\textbf{Effect of Local Zoom Data Augmentation in SFT.}
In SFT, we fine-tune only the MLP layers and ablation different image augmentation strategies. As shown in Table~\ref{tab:zoom_aug_effect}, our model trained with zoom-in patch augmentation achieves an accuracy of 48.54\%, compared to 45.32\% without zoom-in patch.

\begin{table}[htb]
\centering
\begin{tabular}{lc}
\toprule
\textbf{Augmentation} & \textbf{Accuracy (\%)} \\
\midrule
None & 45.32 \\
Zoom-in Patch & \textbf{48.54} \\
\bottomrule
\end{tabular}
\caption{Effect of Zoom-in Augmentation.}
\label{tab:zoom_aug_effect}
\vspace{-3mm}
\end{table}

\noindent\textbf{Effect of Reward Function Components.}
We retain the correctness reward as the primary optimization objective and conduct ablation studies on auxiliary components. Table~\ref{tab:reward_ablation} summarizes the results, which demonstrate that incorporating both format and validity rewards improves performance over SFT alone or RL without SFT. These findings highlight the importance of a comprehensive reward design in guiding the model toward structured, valid, and accurate predictions.

\begin{table}[htb]
\centering
\resizebox{1\linewidth}{!}
{
\begin{tabular}{c c c c c}
\hline
\multirow{2}{*}{\textbf{SFT}} & \multicolumn{3}{c}{\textbf{Reward}} & \multirow{2}{*}{\textbf{Accuracy(\%)}}\\
\cline{2-4}
& \textbf{Correctness} & \textbf{Format} & \textbf{Validity} &  \\
\hline
\checkmark &  &  &  & 48.54 \\
     & \checkmark & \checkmark & \checkmark & 46.95 \\
\checkmark & \checkmark & \checkmark &      & 50.86 \\
\checkmark & \checkmark &      & \checkmark & 51.22 \\
\checkmark & \checkmark & \checkmark & \checkmark & \textbf{52.18} \\
\hline
\end{tabular}
}
\vspace{-3mm}
\caption{Ablation on different reward settings.}
\label{tab:reward_ablation}
\vspace{-3mm}
\end{table}

\begin{figure}[htbp]
    \centering
    \vspace{-2mm}
    \includegraphics[width=0.9\linewidth]{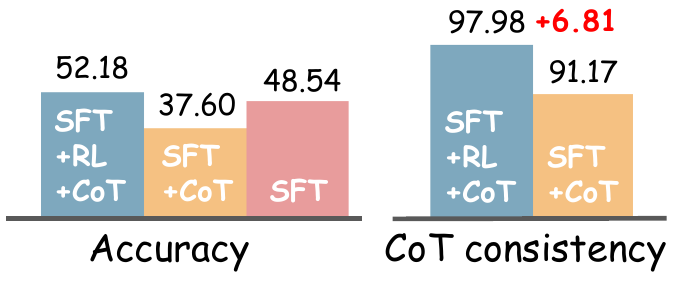} 
    \vspace{-5mm}
    \caption{Accuracy (left) and CoT reasoning–answer consistency (right) under different training and inference setups.}
    \label{fig:cot_results}
    \vspace{-3mm}
\end{figure}

\noindent\textbf{Early-Stage Chain-of-Thought (CoT) Forcing and Consistency Evaluation.}
As shown in Figure~\ref{fig:cot_results}, using first-stage weights, we investigate the effect of forcing CoT generation. Accuracy drops from 48.54\% (without CoT) to 37.60\% (with CoT), suggesting that premature CoT disrupts stable reasoning and distorts the data distribution. Beyond accuracy, we further evaluate reasoning–answer consistency by feeding the generated think part into gpt-oss-20b~\cite{agarwal2025gpt} to predict the most probable class, and comparing this with the model’s final output. The SFT+RL model achieves a consistency of 97.98\%, substantially higher than the purely fine-tuned model (91.17\%). These findings highlight the necessity of the RL stage: only after the model acquires basic diagnostic ability can CoT be effectively induced to yield coherent reasoning while maximizing both consistency and performance. 

\section{CONCLUSION}
\label{sec:typestyle}

In this work, we construct a slice-level CT dataset, CT-RATE-VQA, and propose a zoom-in patch embedding-based data augmentation method to emphasize local lesion features. We apply the GRPO reinforcement learning framework to medical visual question answering and design a composite reward function that integrates structured reasoning, output validity, and prediction correctness. Experimental results demonstrate that our approach improves both disease classification and reasoning capabilities, providing a practical data and training solution for the medical community.







\section{Compliance with ethical standards}
\label{sec:ethics}
Informed consent was obtained from all individual participants involved in the study.

\section{Acknowledgments}
\label{sec:acknowledgments}
Supported by Natural Science Foundation of China under Grant 62271465, National Key R\&D Program of China under Grant 2025YFC3408300, and Suzhou Basic Research Program under Grant SYG202338.

\bibliographystyle{IEEEbib}
\bibliography{strings}

\end{document}